\documentclass[sigconf,nonacm]{acmart}

\usepackage{pifont}
\usepackage{multirow}
\usepackage{xspace}
\usepackage{xurl}
\setcopyright{none}
\renewcommand\footnotetextcopyrightpermission[1]{}

\definecolor{fgnavy}{HTML}{293462}
\definecolor{fgred}{HTML}{D61C4E}
\definecolor{fgamber}{HTML}{FEB139}
\definecolor{fgsteel}{HTML}{8C93AE}
\definecolor{fgbrown}{HTML}{A8701A}
\definecolor{fglight}{HTML}{D9DCE6}
\definecolor{fgbase}{HTML}{F5F6F9}
\newcommand{\fg}{\textsc{ForecastGym}\xspace}

\begin{document}

\title{\fg: A Live, Controlled, and Replayable Benchmark for Forecasting Agents}

\author{Yicheng Tao}
\authornote{Equal contribution.}
\email{yichengtao@cmu.edu}
\affiliation{\institution{Carnegie Mellon University}\city{Pittsburgh}\state{Pennsylvania}\country{USA}}
\author{Yuante Li}
\authornotemark[1]
\email{yuantel@cs.cmu.edu}
\affiliation{\institution{Carnegie Mellon University}\city{Pittsburgh}\state{Pennsylvania}\country{USA}}
\author{Zipei Fan}
\email{fanzipei@jlu.edu.cn}
\affiliation{\institution{Jilin University}\city{Changchun}\country{China}}
\renewcommand{\shortauthors}{Tao, Li, and Fan}

\begin{abstract}
Agents built on large language models (LLMs) increasingly forecast real-world events by searching the Web and reasoning over what they find.
How well such an agent forecasts depends on both the retrieved evidence and the model, making control over that evidence essential for evaluation.
Live evaluations can mix retrieval with reasoning when each system searches on its own, while retrospective search can leak later information through date filters or today's ranking of historical pages.
Comparing models, search sources, and memory strategies requires both fixed evidence and a record of how it evolved before resolution.
We introduce \fg, which records what four commercial web search APIs return for open prediction-market questions every two days and replays this frozen evidence offline once outcomes are known, comparing models, memory, and captured search sources under a fixed evidence-acquisition policy.
Its first release covers 150 questions, 614 evidence states, one per question and collection time, and 131k search results.
Rerunning the same queries after settlement returns results that all but state the outcome for up to 92\% of the questions, despite the tested date filters.
On the frozen evidence, all tested LLMs trail the market in forecasting accuracy.
Our diagnostic analysis suggests that missed public information contributes more to the gap in high-gap states than misreading does.
Switching the search API changes a model's Brier score by up to 0.043, more than the spread across models, while persistent notes score similarly to rereading all evidence at half the input tokens.
The evidence a forecaster sees can matter as much as the model that reads it.
\end{abstract}

\maketitle
\pagestyle{plain}
\hypersetup{pdfauthor={Yicheng Tao, Yuante Li, Zipei Fan},pdftitle={ForecastGym: A Live, Controlled, and Replayable Benchmark for Forecasting Agents},pdfsubject={Forecasting agents and controlled evidence replay},pdfkeywords={forecasting, benchmark, web search APIs, retrieval-augmented LLMs}}

\section{Introduction}
\label{sec:intro}

Forecasting tournaments have shown that trained forecasters and aggregated crowds can predict world events well \citep{mellers2014psychological,tetlock2015superforecasting}, and online prediction markets such as Polymarket\footnote{\url{https://polymarket.com}} and Kalshi\footnote{\url{https://kalshi.com}} let participants trade contracts on such events, with prices that move as news appears on the Web.
Large language model (LLM) agents now take on a similar role~\citep{hsieh2024reasoning,yuan2025evolvecast}, searching the Web, reasoning over what they find, and revising their probability as new information arrives.
Augmented with retrieval \citep{lewis2020retrieval}, such systems approach the accuracy of human forecasting crowds \citep{halawi2024forecasting,schoenegger2024silicon}, which calls for evaluating them under comparable conditions.
Crucially, a forecast depends on both the \textit{evidence} behind it and the \textit{model} that reads it.
Existing benchmarks follow two routes: \textbf{\ding{182} live} benchmarks collect forecasts on open questions and score them after resolution \citep{karger2025forecastbench,zeng2025futurex,yang2025prophetarena}, and \textbf{\ding{183} retrospective} benchmarks replay resolved questions with historical evidence \citep{zou2022autocast,wildman2025btf,ye2026hindcast,goel2026futuresim}, summarized in Table~\ref{tab:benchmark-comparison}.

Despite these advances, evaluating forecasting agents still faces four challenges.
(i)~\textbf{Leaky reconstruction:} retrospective evaluation must recover the evidence available at time $t$.
Date-filtered retrieval can admit content updated after $t$ \citep{paleka2025pitfalls,ellahib2026temporal}, and even historical page snapshots can be selected or ranked using today's search index \citep{wildman2025btf,prophet2026searchbench}.
Preserving old content alone therefore does not preserve what a search service would have returned then.
(ii)~\textbf{Entangled retrieval and reasoning:} when live benchmarks let each system gather its own evidence, a score mixes what it found with how it reasoned \citep{tang2026drcik,chi2026worldreasoner}.
Shared contexts can isolate model differences \citep{yang2025prophetarena}, but reproducible comparisons also require keeping the evidence fixed across later runs.
(iii)~\textbf{Uncontrolled evidence dynamics:} evidence evolves as an event approaches, and how agents revise their beliefs as it does \citep{yuan2025evolvecast} can be compared only if they experience the same evidence trajectory.
Repeated live forecasts alone do not supply a reusable trajectory on which to vary memory while holding incoming evidence fixed.
(iv)~\textbf{Search-source variation:} the search provider changes what an agent sees, so its effect must be separated from that of the model.
Provider comparisons on static factual tasks \citep{soni2025versa,selvam2026unequal} and time-gated search tests on resolved questions \citep{prophet2026searchbench} motivate a further requirement: comparing sources alongside models and memory on the same prospectively recorded forecasting trajectories.

\begin{figure}[t]
\centering
\includegraphics[width=\columnwidth]{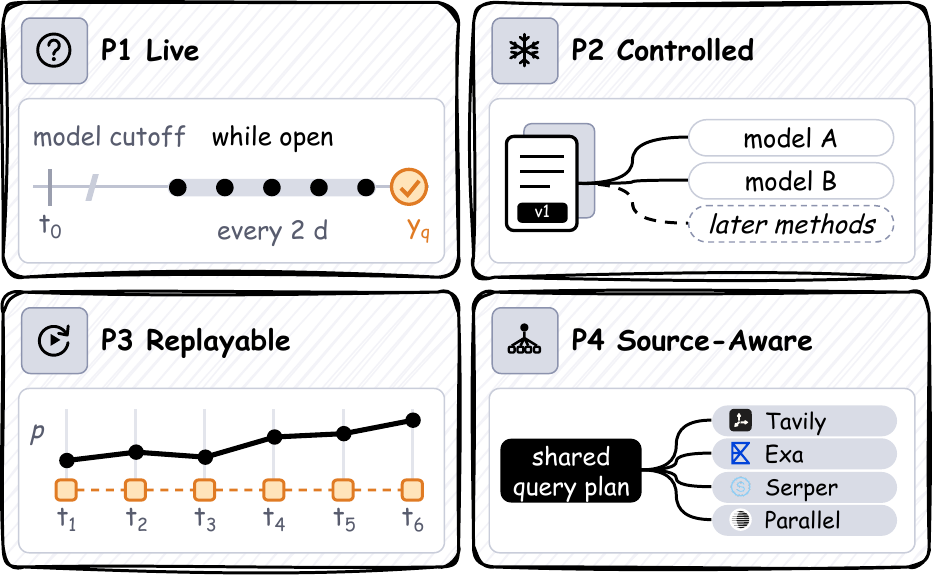}
\caption{Design principles of \fg.}
\Description{Four design properties: unresolved questions at collection, the same time-valid corpus for all methods, replayable evidence trajectories, and separate search providers under a shared initial query plan.}
\label{fig:properties}
\end{figure}

\begin{table}[t]
\caption{Forecasting benchmarks by the principles of Figure~\ref{fig:properties}. Live: open questions at collection. Controlled: reusable fixed evidence. Replayable: multiple states and controlled memory studies. Source-aware: controlled provider comparisons. $\circ$: partial.}
\label{tab:benchmark-comparison}
\centering
\footnotesize
\setlength{\tabcolsep}{2.5pt}
\begin{tabular}{@{}lccccc@{}}
\toprule
\multirow{2}{*}{\textbf{Benchmark}} & \textbf{Live} & \textbf{Controlled} & \multicolumn{2}{c}{\textbf{Replayable}} & \textbf{Source-aware} \\
\cmidrule(lr){2-2}\cmidrule(lr){3-3}\cmidrule(lr){4-5}\cmidrule(l){6-6}
 & Open & Reusable & Multi-state & Memory & Providers \\
\midrule
ForecastQA~\cite{jin2021forecastqa} & -- & \checkmark & -- & -- & -- \\
ForecastBench~\cite{karger2025forecastbench} & \checkmark & -- & -- & -- & -- \\
FutureX~\cite{zeng2025futurex} & \checkmark & -- & -- & -- & -- \\
Prophet Arena~\cite{yang2025prophetarena,prophet2026searchbench} & \checkmark & $\circ$ & \checkmark & -- & \checkmark \\
BTF~\cite{wildman2025btf} & -- & \checkmark & -- & -- & -- \\
Hindcast~\cite{ye2026hindcast} & -- & \checkmark & -- & -- & -- \\
FutureSim~\cite{goel2026futuresim} & -- & \checkmark & \checkmark & \checkmark & -- \\
\textbf{\fg} & \checkmark & \checkmark & \checkmark & \checkmark & \checkmark \\
\bottomrule
\end{tabular}
\end{table}

\begin{figure*}[t]
\centering
\includegraphics[width=0.94\textwidth]{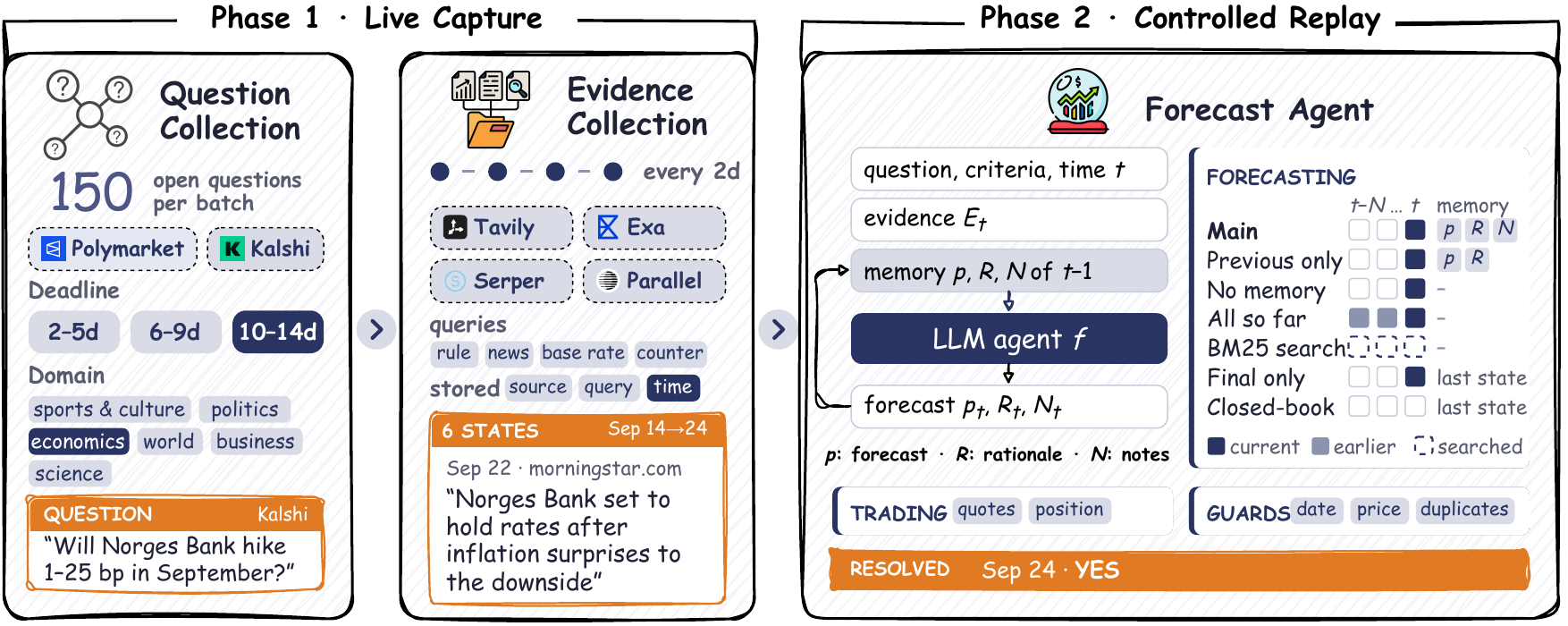}
\caption{The \fg lifecycle described in Section~\ref{sec:benchmark}. Phase~1 records what four search APIs return every two days for open questions, with each result stored with its source, query, and retrieval time. Phase~2 replays the frozen states to a forecasting agent behind three safeguards and changes one argument of Eq.~(\ref{eq:agent}) at a time. Orange cards follow one question of the first batch.}
\label{fig:lifecycle}
\end{figure*}

To address these challenges, we introduce \textbf{\fg}, a benchmark that treats evidence as a controlled experimental variable, as illustrated in Figure~\ref{fig:properties}.
Its key insight is simple: \emph{what a search API returns while a question is still open was knowable at that time by construction.}
\fg enables retrospective experimentation under prospectively recorded information conditions, preserving both the evidence content and the search provider's selection before resolution.
Figure~\ref{fig:lifecycle} shows its two phases.
In \textbf{Phase~1, Live Capture}, \fg queries four commercial web search APIs every two days for unresolved prediction-market questions and stores each provider's results as a time-stamped evidence state, as detailed in Section~\ref{sec:capture}.
Providers share an initial query plan and follow-up policy, and their results are stored separately.
In \textbf{Phase~2, Controlled Replay}, agents follow the protocol in Section~\ref{sec:replay} to read the frozen states offline in chronological order once outcomes are known.
Replay isolates forecasting and belief revision: for a given provider, every run follows the same evidence trajectory.
The \emph{model}, \emph{memory}, and \emph{search source} can then be varied one at a time while the other settings remain fixed, allowing score differences to be attributed to the component changed.

The first release contains 150 questions, 614 evidence states, each holding one question's evidence at one collection time, and 131k search results.
The same queries rerun after settlement reveal the outcome for up to 92\% of the questions despite the tested date filters.
On the frozen evidence, all tested LLMs trail the market in forecasting accuracy.
Our diagnosis of high-gap states attributes nearly three times as much of their gap to missed public information as to misreading.
Switching the search API moves the Brier score by up to 0.043, more than the spread across models, while memory mainly steadies the forecasts.
\emph{The evidence a forecaster sees can matter as much as the model that reads it.}

Our contributions are:
(1)~\emph{Live: a reusable collection protocol and pipeline}, which records search results while questions are unresolved and demonstrates leakage under the tested post-hoc date filters;
(2)~\emph{Controlled: an evaluation protocol on fixed evidence}, which separates model comparison from evidence acquisition and establishes model and market baselines;
(3)~\emph{Replayable: recorded evidence trajectories and sequential replay}, which support controlled memory experiments and show that memory mainly steadies forecasts;
and (4)~\emph{Source-Aware: a controlled comparison of four commercial search APIs}, showing that source differences can exceed model differences in accuracy.

\section{Related Work}

\paragraph{Forecasting benchmarks.}
Early datasets framed event forecasting over time-restricted text: ForecastQA~\cite{jin2021forecastqa} casts it as question answering, and Autocast~\cite{zou2022autocast} pairs forecasting-tournament questions with a date-indexed news corpus.
Subsequent systems retrieve news, reason with tools, or aggregate several models to produce probabilistic forecasts \citep{halawi2024forecasting,hsieh2024reasoning,schoenegger2024silicon}.
Live benchmarks, including ForecastBench~\cite{karger2025forecastbench}, FutureX~\cite{zeng2025futurex}, and Prophet Arena~\cite{yang2025prophetarena}, evaluate predictions submitted before outcomes are known, and refresh their questions to avoid contamination, as LiveBench~\cite{white2025livebench} does for general tasks.
Along the same line, Daily Oracle~\cite{dai2025prescient} generates questions continuously from daily news, and MIRAI~\cite{ye2024mirai} evaluates agents that query event databases.
Other work scores forecasters before resolution by testing whether their predictions on related questions agree~\cite{paleka2025consistency}, or scores financial forecasts under trading costs~\cite{hu2026fintsb}.
Retrospective benchmarks instead provide reusable historical environments: Bench to the Future~\cite{wildman2025btf} supplies offline web corpora, and Hindcast~\cite{ye2026hindcast} replays prediction-market questions against a frozen Reddit archive.
Such evaluations must account for both model knowledge of resolved outcomes and the temporal validity of retrieved evidence \citep{paleka2025pitfalls,ellahib2026temporal}.
Live evaluation does not by itself provide fixed evidence for later controlled comparisons, while retrospective evaluation must preserve both historical content and the search service's selection of it.
For example, BTF freezes page content but uses live Google search over its stored pages, leaving a possible change in ranking \citep{wildman2025btf}.
\fg combines prospective collection with offline replay, preserving the results exactly as each provider returned them before resolution.

\paragraph{Search evidence for agents.}
Because the knowledge of language models goes stale~\cite{lazaridou2021mind,liska2022streamingqa}, questions about the current world are answered better with search, as RealTime QA~\cite{kasai2023realtimeqa} and FreshLLMs~\cite{vu2024freshllms} show, which makes the retrieved evidence part of what is evaluated.
Search benchmarks therefore fix it: BrowseComp-Plus~\cite{chen2026browsecompplus} replaces live search with a curated corpus to separate retrieval from reasoning, and Deep Research Bench~\cite{bosse2025drb} freezes scraped pages so that agents can be compared offline.
Within forecasting, Dr-CiK~\cite{tang2026drcik} tests retrieval of supporting context for time-series predictions, and WorldReasoner~\cite{chi2026worldreasoner} scores the evidence an event forecast cites.
The search provider adds its own variation: OpenHands-Versa~\cite{soni2025versa} succeeds on GAIA at different rates with different providers, and Selvam and Ghosh~\cite{selvam2026unequal} find similar accuracy but unequal evidence across providers on SealQA-Hard.
Prophet Arena audits time-gated search with prospective captures, but acknowledges ranking leakage and fallible leakage detection \citep{prophet2026searchbench}.
Its exports omit page text, precluding exact offline replay from those exports.
Among the benchmarks in Table~\ref{tab:benchmark-comparison}, \fg alone combines prospectively captured, provider-specific evidence trajectories with controlled offline comparisons of models and memory.

\paragraph{Belief revision and memory.}
EvolveCast~\cite{yuan2025evolvecast} compares forecasts revised after news updates with human revisions, for both individual and accumulated news, although its question--news associations are reconstructed retrospectively.
FutureSim~\cite{goel2026futuresim} replays a chronological news stream to study long-horizon adaptation, search, and memory.
ForecastCompass~\cite{chang2026forecastcompass} instead transfers experience across forecasting tasks through adaptive factor and reasoning memories, while LongMemEval~\cite{wu2025longmemeval} examines retention and knowledge updates in extended conversations.
Yet the forecasting studies among them either rebuild the evidence after the fact or transfer memory across tasks, so none isolates memory on evidence captured before resolution.
\fg addresses this gap by holding each provider's evidence trajectory for the same question fixed and varying only what the agent retains, so that differences in forecasts can be attributed to memory alone.

\section{The \fg Benchmark}
\label{sec:benchmark}

\fg is built to answer one question: \textit{\textbf{how does a forecast change when a single component of the forecaster changes?}}
Section~\ref{sec:formulation} formalizes this question and the requirements it places on the evidence, and the remaining subsections show how the two phases of Figure~\ref{fig:lifecycle} meet them.

\subsection{Problem Formulation}
\label{sec:formulation}

A forecasting agent that searches the Web combines three components: a \textit{search source} that acquires evidence, a \textit{memory} that carries it across time, and a \textit{model} that reasons over it.
At time $t$, an agent with search provider $k$ forecasts a binary question $q$ as
\begin{equation}
p_t=f\bigl(q,\ E^{(k)}_t,\ M_{t-1}\bigr),
\label{eq:agent}
\end{equation}
where $E^{(k)}_t$ is the evidence that provider $k$ returns at time $t$, $M_{t-1}$ what the agent carries from earlier times, and $f$ the model.

\begin{definition}[Evidence state]
Let $\mathcal{K}$ be the set of search providers.
The pair $(q,t)$, with $t$ in the set $\mathcal{T}_q$ of times at which evidence for $q$ is collected, is a \emph{state}, to which each provider $k\in\mathcal{K}$ contributes the evidence $E^{(k)}_t$.
A benchmark instance is the tuple
\begin{equation}
\bigl(q,\ \mathcal{T}_q,\ \{E^{(k)}_t\}_{t\in\mathcal{T}_q,\,k\in\mathcal{K}},\ y_q\bigr),
\label{eq:instance}
\end{equation}
in which the outcome $y_q\in\{0,1\}$ is added only after resolution.
\end{definition}

\fg evaluates forecasters through controlled comparisons, each of which changes a single argument of Eq.~(\ref{eq:agent}).
Let $\mathcal{L}(f,k,M)$ be the score of an agent on a batch, such as its mean Brier score over questions as defined in Section~\ref{sec:setup}.
To compare models $f$ and $f'$ while holding the search source and memory strategy fixed, we measure
\begin{equation}
\Delta_{f\to f'}=\mathcal{L}(f',k,M)-\mathcal{L}(f,k,M),
\label{eq:effect}
\end{equation}
and the effects $\Delta_{k\to k'}$ of the search source and $\Delta_{M\to M'}$ of the memory are defined in the same way.

Estimating Eq.~(\ref{eq:effect}) places three requirements on the evidence.
\textbf{R1. Time validity:} every item in $E^{(k)}_t$ was knowable at time $t$, since otherwise the effect is measured on leaked answers.
\textbf{R2. Source isolation:} provider comparisons fix collection times, initial queries, and the follow-up rule and budget.
\textbf{R3. Invariance across runs:} both terms of Eq.~(\ref{eq:effect}) read the same evidence.
Live evaluation does not guarantee R3, while retrospective search must guard R1 against later content and ranking signals.
Sections~\ref{sec:validity} and~\ref{sec:capture} show how Live Capture meets R1 and R2; Controlled Replay meets R3 by fixing the evidence across runs, as described in Section~\ref{sec:replay}.

\subsection{Temporal Validity by Construction}
\label{sec:validity}

Writing $\mathcal{W}_t$ for the content available on the Web up to time $t$, requirement R1 asks that $E_t\subseteq\mathcal{W}_t$.
Retrospective benchmarks search after resolution, at some time $T>t$, and keep the results dated before $t$,
\begin{equation}
E^{\mathrm{date}}_t=\bigl\{d\in S(Q,\mathcal{W}_T) : \mathrm{date}(d)\leq t\bigr\},
\label{eq:datefilter}
\end{equation}
where $S(Q,\mathcal{W})$ denotes the results of queries $Q$ on the Web $\mathcal{W}$.
This set errs in both directions.
A page without a date is dropped even if it was knowable at $t$, and a page whose content was replaced after $t$, such as a game preview later rewritten into a box score, keeps its early date and passes the filter with the outcome inside.
\fg instead issues the queries $Q_t$ at time $t$ itself and records
\begin{equation}
E^{(k)}_t=S_k(Q_t,\mathcal{W}_t)\subseteq\mathcal{W}_t,
\label{eq:capture}
\end{equation}
so R1 holds by construction, whatever dates the pages carry.

Retrieval time thus replaces publication date as the bound on what was knowable, and Sections~\ref{sec:stats} and~\ref{sec:leakage} measure how far Eq.~(\ref{eq:datefilter}) and Eq.~(\ref{eq:capture}) diverge in practice.
Live capture also fixes the provider's selection at $t$: even a search over historical page snapshots can otherwise use a later index or ranking \citep{wildman2025btf,prophet2026searchbench}.
Validity concerns only when an item became knowable, not whether it is safe to show, so Controlled Replay adds safeguards.

\subsection{Live Capture}
\label{sec:capture}

We use Polymarket and Kalshi to ground evaluation in real prediction markets, with tradable questions, observed prices as an external baseline, and platform settlements as outcomes.

\paragraph{Rolling collection protocol.}
Answers that already exist can be memorized, and a fixed question set falls behind every new model.
\fg supplies a common protocol for users to collect fresh binary questions closing within 14 days.
Batch size and domain quotas are configurable; this paper evaluates one budget-limited batch of 150 questions.
A model is evaluated only on batches opened after its release.
To keep questions that are still uncertain, settle unambiguously, and leave time to revise a forecast, we drop markets priced below 10\% or above 90\% or closing within 48 hours, and an LLM reviewer removes questions with vague rules or hard-to-verify outcomes.
Selection then balances horizon, domain, and trading activity, and keeps at most one question per real-world event, following the rules in Appendix~\ref{app:batch}.

\begin{figure*}[t]
\centering
\includegraphics[width=0.94\textwidth]{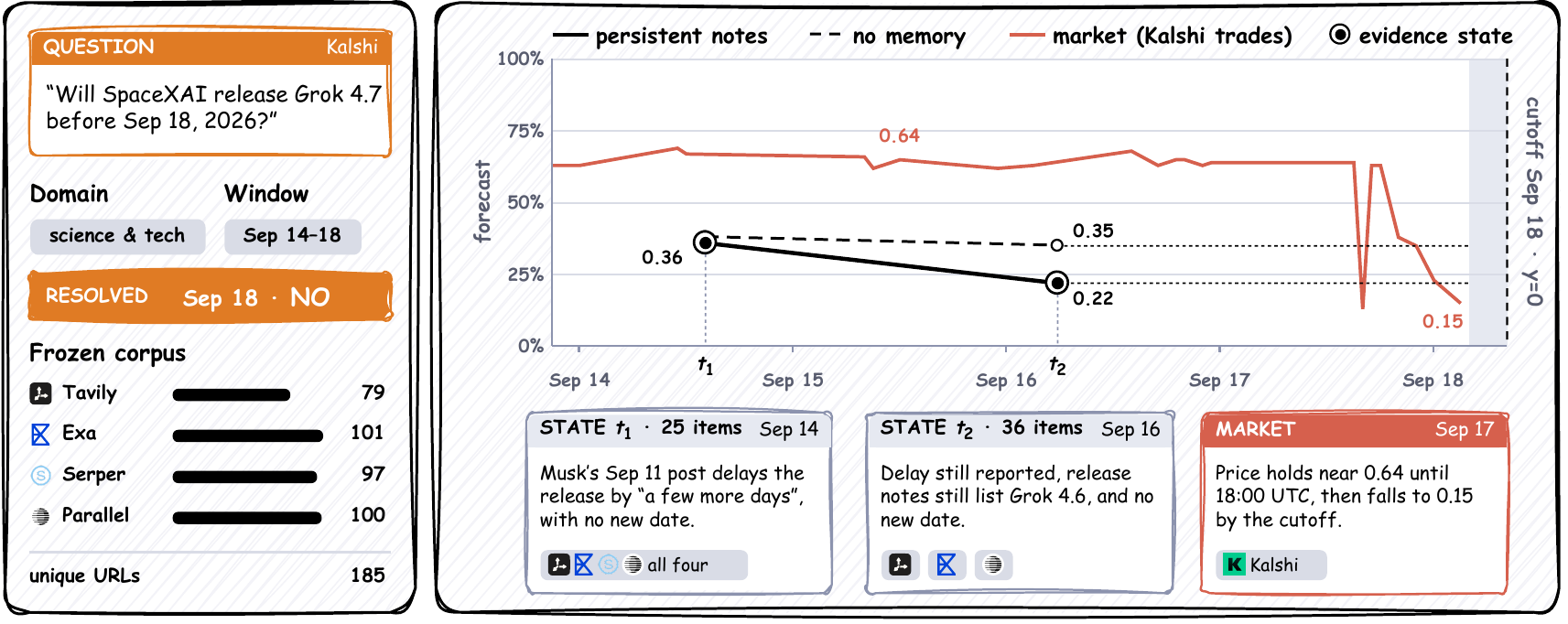}
\caption{One question of the first batch, replayed by Gemini~3.8 Flash on Tavily's evidence. The market price is never shown to the model.}
\Description{Left: the question, its domain, horizon, two evidence states, the outcome No, and per-provider result counts. Right: forecasts over time. With persistent notes Gemini falls from 0.36 to 0.22 and without memory from 0.38 to 0.35, while the Kalshi price stays near 0.64 until 17 September and then falls to 0.15 before the cutoff.}
\label{fig:example}
\end{figure*}

\paragraph{Evidence trajectories.}
\fg collects each question's evidence every two days until the forecast cutoff, the earliest of its deadline, the start of the event, and its settlement, so the states follow the evidence as the event approaches but never report the event itself.
Every result is stored with its source, query, and retrieval time, and a late collection is marked as missed rather than reconstructed, so Eq.~(\ref{eq:capture}) holds for every stored item.

\paragraph{Source comparison under a shared query policy.}
At each collection time \fg queries four web search APIs that stand for different ways of searching.
Serper returns Google's ranked results, Tavily and Parallel are built for LLM agents and return page content or excerpts chosen for the query, and Exa searches its own index by meaning rather than by keywords.\footnote{\url{https://serper.dev}, \url{https://tavily.com}, \url{https://parallel.ai}, \url{https://exa.ai}}
To meet R2, an LLM writes a shared initial plan of up to four queries per collection time.
A resolution query identifies the rule and source that will settle the question; a current-evidence query retrieves the latest news.
A base-rate query asks how often similar events occur, and a counterevidence query seeks reasons the likely outcome may fail.
Where a provider's results miss one of these needs, the LLM sends that provider up to two follow-up queries.
At each state, queries are planned during capture, independently of the forecasting models, so one recording serves every model.

\paragraph{Freezing the record.}
Outcomes are read from the platform that posed the question, and disputed or ambiguous ones are excluded.
A completed batch is frozen as a versioned snapshot, and unsettled questions are scored later without changing their evidence.

\subsection{Controlled Replay}
\label{sec:replay}

Controlled Replay evaluates forecasters on the frozen record offline in Phase~2 of Figure~\ref{fig:lifecycle}, much as a trader backtests a strategy on recorded market data rather than on prices reconstructed afterwards.
The agent sees one provider's states of a question in order, as if it were forecasting live, and forecasts each state once without calling any live service, so every run reads the same evidence and R3 holds.

\paragraph{Replay safeguards.}
A valid state can still give a model a shortcut, so three safeguards filter it before it is shown.
Items dated later than the state, such as previews dated by an upcoming event, are hidden, and so are items that show the question's own market price, which the model could otherwise copy.
Near-duplicates are merged, so that repeated coverage does not count as more evidence, and everything else is shown regardless of search rank.

\paragraph{Memory as an explicit state.}
Following retrieval-augmented forecasters \citep{halawi2024forecasting}, a lightweight agent reads the question, its resolution criteria, the current time, and the evidence $E_t$, and returns a probability $p_t$, the items it relied on, and a short rationale $R_t$.
Since a state shows only what was retrieved at that time, the agent needs memory, and keeping it in a growing context would tie it to the model's input.
\fg instead passes memory between states as an explicit object that can be replaced like the other arguments of Eq.~(\ref{eq:agent}).
In the main condition it holds the previous forecast and rationale and notes $N_{t-1}$ that the agent rewrites at every state, covering the key evidence with its dates, the factors that matter, and how it has been updating, in the spirit of ForecastCompass \citep{chang2026forecastcompass}:
\begin{equation}
M^{\mathrm{main}}_{t-1}=\bigl(p_{t-1},\ R_{t-1},\ N_{t-1}\bigr).
\label{eq:main}
\end{equation}
At the first state, the previous forecast is absent and there are no notes from earlier states.

\paragraph{Single-factor interventions.}
Every comparison in Section~\ref{sec:results} starts from this main condition and changes one argument of Eq.~(\ref{eq:agent}), the model, the provider, the evidence, or the memory, as Table~\ref{tab:variants} lists.
Previous only retains the preceding forecast and rationale but drops the notes; No memory carries none of these forward.
Two memory variants replace memory with all evidence so far, $E_{\leq t}=\bigcup_{\tau\leq t}E_\tau$, read in full or searched with the keyword-based ranking function BM25 \citep{robertson2009bm25}.
Beyond these interventions, a trading mode shows the agent the market's quotes and its own position and lets it buy, hold, or close at each state.

\begin{table}[t]
\caption{Single-factor interventions on Eq.~(\ref{eq:agent}). Each row changes one argument of the main condition and keeps the others as in Main. Evidence: what the agent reads at state $t$. Memory: what it carries from the state before.}
\label{tab:variants}
\centering
\footnotesize
\setlength{\tabcolsep}{3.5pt}
\begin{tabular}{@{}llll@{}}
\toprule
\textbf{Argument} & \textbf{Variant} & \textbf{Evidence} & \textbf{Memory} \\
\midrule
 & Main & $E^{(k)}_t$ & $p_{t-1}, R_{t-1}, N_{t-1}$ \\
\midrule
Model $f$ & Four LLMs & as Main & as Main \\
Source $k$ & Four providers & $E^{(k')}_t$ & as Main \\
\midrule
Evidence $E$ & (i) Closed-book & none, last state & none \\
 & (ii) Final only & $E_t$, last state & none \\
\midrule
Memory $M$ & (iii) Previous only & $E_t$ & $p_{t-1}, R_{t-1}$ \\
 & (iv) No memory & $E_t$ & none \\
 & (v) All so far & $E_{\leq t}$ & none \\
 & (vi) BM25 search & BM25 over $E_{\leq t}$ & none \\
\bottomrule
\end{tabular}
\end{table}

\section{The First \fg Release}
\label{sec:stats}

The first batch opened on 14 September 2026 and was collected for 14 days.
It contains 150 questions, 614 evidence states and 131k search results, with two to seven states per question.
The batch targets forecasting on real prediction-market questions, with domain quotas broadening coverage across six domains within the platforms' supply.
One question was cancelled and five await settlement, leaving 144 scored questions.
Their initial market prices score 0.215 Brier, against 0.250 for a constant forecast of 0.5, consistent with uncertainty at collection.
Appendix~\ref{app:batch} reports the full statistics and scope.

Before evaluating any forecaster, we use the batch to test the three premises behind the design of Section~\ref{sec:benchmark}: that the search source shapes the evidence, that the evidence evolves, and that publication dates cannot bound what was knowable.

\begin{figure}[t]
\centering
\includegraphics[width=0.94\columnwidth]{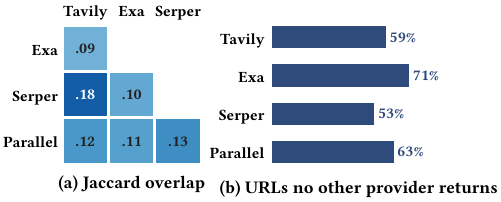}
\caption{The four providers return largely different pages under the shared query policy, averaged over all states. Jaccard overlap: shared URLs over all URLs of the pair.}
\Description{Heat map of pairwise Jaccard overlaps between providers, from 0.09 to 0.18, and bars showing that 53 to 71 percent of each provider's URLs come from no other provider.}
\label{fig:overlap}
\end{figure}

\paragraph{Search providers largely disagree.}
As Figure~\ref{fig:overlap} shows, for the same question and time any two providers' URL sets overlap little, and four in five pooled URLs come from exactly one provider.
Even under this shared policy, provider choice changes most of the returned URLs.

\paragraph{Evidence keeps changing.}
At a question's second state, 71\% of the pooled URLs are new, and at every later state more than 40\% still are.
Over its lifetime a question accumulates 374 distinct URLs on average, about three times as many as a single state holds, so a single snapshot would miss much of what became knowable.

\paragraph{Page dates are unreliable.}
The date filter of Eq.~(\ref{eq:datefilter}) errs in two directions, and the batch shows that both are common.
Most pages carry no date and would be dropped: only 36.8\% of our documents have one.
Some dates are also wrong: 6.7\% of dated documents are dated after they were retrieved, typically previews and box-score pages dated by the upcoming event, whose content is later replaced by the outcome.
This matches the leakage reported for date-filtered retrieval \citep{ellahib2026temporal}, whereas Eq.~(\ref{eq:capture}) bounds every document by its retrieval time.

\section{Experimental Setup}
\label{sec:setup}

Following Figure~\ref{fig:properties}, RQ1 tests \emph{Live} capture; RQ2 uses \emph{Controlled} evidence to compare models and diagnose errors; RQ3 studies memory and belief revision through \emph{Replayable} trajectories; and RQ4 tests \emph{Source-Aware} provider comparisons.

\paragraph{RQ1: Can post-hoc search replace live capture?}
We sample 60 settled questions in proportion to the batch's domains.
For each question's last replay state $t$, we rerun its two initial queries for current evidence and counterevidence after settlement, once raw and once with each provider's filter for pages published before $t$.
An outcome-aware LLM judge marks an item as leaking if its content carries information from after $t$, whatever date the page shows.
We report Leaking, the share of leaking items, and Revealed, the share of questions whose leaking items all but state the outcome.
Gemini~3.8 Flash forecasts from the filtered and live results of the same queries; Appendix~\ref{app:leakage} gives the filter settings.

\paragraph{RQ2: How do models compare with the market, and why?}
All four models forecast every state in the main condition with Tavily's evidence and once more closed-book.
We compare their scores with the market price at the same states and when collection began.
Under the \textbf{blind rule}, we buy \$1 of the side a model favors whenever its forecast differs from the quote by more than 0.05; in the \textbf{trading mode}, Gemini~3.8 Flash makes its own trades with the quotes in view.
Trades fill at the recorded price plus or minus half the bid--ask spread, pay Kalshi's fee, and are scored by profit per dollar staked, with details in Appendix~\ref{app:trading-sim}.
We use outcome-aware Gemini~3.8 Flash for two distinct diagnostic labeling tasks.
Evidence-direction labeling classifies each state's evidence as pointing toward the outcome, against it, or both ways.
For states trailing the market by more than 0.1 Brier, gap attribution identifies the fact that would have improved the forecast and checks the displayed evidence, items removed by the safeguards, and other providers' results, as detailed in Appendix~\ref{app:labels}.

\paragraph{RQ3: Do forecasts sharpen, and does memory help?}
We track scores by days until the event and ablate memory for Gemini~3.8 Flash with Tavily's evidence.
For each pair of consecutive states, we compare the forecast's absolute probability change and direction with the market's move over the same interval, as detailed in Appendix~\ref{app:memory}.

\paragraph{RQ4: How much does the search source matter?}
Gemini~3.8 Flash replays every state with each provider's evidence.
To see whether the provider also reorders the models, all four forecast the last state without memory from Tavily's evidence and from Exa's, whose results overlap least with Tavily's.

\paragraph{Models.}
We evaluate four LLMs from four developers, GPT-5.6 Terra from OpenAI, Claude Sonnet~5 from Anthropic, Gemini~3.8 Flash from Google, and the open-weight DeepSeek-V4.1-Flash from DeepSeek-AI~\cite{deepseek2026v41flash}.
All were released before the batch opened, as listed in Appendix~\ref{app:models}, so none can have seen its outcomes.
All receive the same prompt and settings given in Appendix~\ref{app:prompt}.
Gemini~3.8 Flash, chosen for its speed and cost, runs the variants beyond the main comparison.
GPT-6.1 Sol, released after the batch resolved, is not ranked but checks whether a newer model closes the gap and whether knowledge of the outcomes shows in its closed-book forecasts.

\paragraph{Choice of providers.}
Tavily is the main provider, chosen before any forecast from properties measured without outcomes.
At collection time, the query-planning LLM rated its results highest for covering the query purposes, and they held the largest share of items relevant to the question, although it reports no publication dates, as Appendix~\ref{app:providers} details.

\paragraph{Metrics.}
Our primary metric is the Brier score $(p-y_q)^2$, the squared error of a forecast probability.
It is strictly proper, so a forecaster scores best in expectation by reporting its true belief \citep{brier1950,gneiting2007proper}, and a constant forecast of 0.5 scores 0.25.
To score a whole trajectory, the time-averaged Brier score (TBS) averages it over the $T_q$ scored states of a settled question $q$,
\begin{equation}
\mathrm{TBS}_q=\frac{1}{T_q}\sum_{t}\bigl(p_{q,t}-y_q\bigr)^2,
\label{eq:tbs}
\end{equation}
and the mean TBS or the mean last-state Brier score over questions serves as the score $\mathcal{L}$ of Eq.~(\ref{eq:effect}).
The last state is the latest before the forecast cutoff.
Accuracy is the share of last-state forecasts on the correct side of 0.5, counting a forecast of exactly 0.5 as half.
The market price at each state, the platform's latest price at or before the replay time, serves as an external baseline, since traders are not limited to the recorded evidence.
Differences are paired by question, averaging over a question's states first, and come with 95\% bootstrap intervals obtained by resampling questions.
We call a difference significant when its interval excludes zero, and compare model rankings with Kendall's $\tau$.

\section{Results and Analysis}
\label{sec:results}

Sections~\ref{sec:leakage}, \ref{sec:controlled-results}, \ref{sec:memory-results}, and~\ref{sec:source-results} address RQ1--RQ4, respectively.

\begin{table}[t]
\caption{Post-hoc search on 60 settled questions, with raw and date-filtered results. $^*$Significantly below the live snapshot.}
\label{tab:leakage}
\centering
\footnotesize
\setlength{\tabcolsep}{3.2pt}
\begin{tabular}{@{}lccccc@{}}
\toprule
 & \multicolumn{2}{c}{\textbf{Leaking}} & \multicolumn{2}{c}{\textbf{Revealed}} & \textbf{Brier} \\
\cmidrule(lr){2-3}\cmidrule(lr){4-5}\cmidrule(l){6-6}
\textbf{Results} & raw & filtered & raw & filtered & filtered \\
\midrule
\textbf{Live snapshot, Tavily}\textsuperscript{※} & \multicolumn{2}{c}{\textbf{3\%}} & \multicolumn{2}{c}{\textbf{10\%}} & \textbf{0.224} \\
\midrule
Tavily & 42\% & 18\% & 87\% & 60\% & 0.137$^*$ \\
Exa & 31\% & 27\% & 90\% & 92\% & 0.080$^*$ \\
Serper & 32\% & 17\% & 73\% & 43\% & 0.194 \\
Parallel & 37\% & n/a & 90\% & n/a & -- \\
\bottomrule
\end{tabular}
\par\smallskip
\parbox{\columnwidth}{\footnotesize ※The live snapshot is not at zero because its flagged cases are mostly results already public at $t$, such as a review score already above the threshold, or LLM-judge false alarms, such as an article about an earlier game.\par
Parallel has no filtered run: its filter only keeps pages published after a date.}
\end{table}

\subsection{Post-hoc Search Leaks the Outcome}
\label{sec:leakage}

\paragraph{Date filters reduce leakage but do not stop it.}
As Table~\ref{tab:leakage} shows, the date-filtered evidence of Eq.~(\ref{eq:datefilter}) still leaks for every provider, and each fails in its own way.
Tavily attaches no dates to its results, so its filter cannot be audited at all.
Exa's filter removes every result it dates after $t$ but passes the 61\% of results it leaves undated, and Serper's still returns 12\% of results dated after $t$.
The leaking items are mostly final scores, earnings releases, and product changelogs, while the judge flags few live results, providing a check on false alarms.

\paragraph{The leaks inflate the scores.}
Forecasting from Tavily's or Exa's filtered results, Gemini scores 0.137 and 0.080 Brier, far better than any model or even the market does on live evidence in Table~\ref{tab:main}.
The tested date filters thus leave post-hoc evidence temporally unreliable, motivating capture before resolution, as in the remaining experiments.

\subsection{Models Trail the Market}
\label{sec:controlled-results}

\subsubsection{Model Differences Are Small}
\label{sec:model-comparison}

\begin{table}[t]
\caption{Main condition with Tavily's evidence, at each question's last state. Gap: to the market over all states, significant for all. $^\dagger$Released after the batch resolved.}
\label{tab:main}
\centering
\footnotesize
\setlength{\tabcolsep}{3pt}
\begin{tabular}{@{}lcccc@{}}
\toprule
\textbf{Forecaster} & \textbf{Last-state Brier}$\downarrow$ & \textbf{Accuracy}$\uparrow$ & \textbf{Closed-book}$\downarrow$ & \textbf{Gap} \\
\midrule
GPT-5.6 Terra & \textbf{0.217} & 0.642 & 0.260 & \textbf{$+$0.022} \\
Claude Sonnet 5 & 0.244 & 0.597 & 0.279 & $+$0.046 \\
Gemini 3.8 Flash & 0.218 & 0.628 & 0.300 & $+$0.028 \\
DeepSeek-V4.1-Flash & 0.220 & \textbf{0.660} & \textbf{0.258} & $+$0.028 \\
\midrule
GPT-6.1 Sol$^\dagger$ & 0.208 & 0.653 & 0.266 & $+$0.018 \\
\midrule
\textit{Market} & 0.176 & 0.719 & & \\
\bottomrule
\end{tabular}
\end{table}

Table~\ref{tab:main} gives the results, and the closed-book scores show what the evidence adds, which can be negative \citep{tang2026drcik,ye2026hindcast}.

\paragraph{On the same evidence, the models are hard to tell apart.}
Score differences among GPT-5.6 Terra, Gemini~3.8 Flash, and DeepSeek-V4.1-Flash are not significant, and accuracy orders them differently than the Brier score does.
Only Claude Sonnet~5 falls clearly behind, through overconfidence at the low end: its forecasts near 0.05 came true 38\% of the time, and in Figure~\ref{fig:cases}c it took undated pages about a 2018 win for the upcoming match.
All models lean on the newest evidence, as 92--95\% of the items they cite were first retrieved at the latest state.
GPT-6.1 Sol, released after every question had settled, is within noise of the top three, and its closed-book score shows no sign of knowing the outcomes.

\paragraph{The evidence is what makes these forecasts useful.}
Without it, no model beats a constant forecast of 0.5, so background knowledge does not suffice, and with it every model improves, significantly so for GPT-5.6 Terra and Gemini.

\paragraph{Yet no model beats the market.}
Every model trails the market at the last state and over the full trajectory.
Excluding the 216 states priced at a bid--ask midpoint leaves the gap unchanged.
Nor does combining models close it: their mean is no better than the best single model, and even a blend weighted toward the market falls short of the market alone, as shown in Appendix~\ref{app:ensemble}.

\paragraph{The tested trading strategies yield no significant profit.}
A forecaster that trails the market could still profit if it knew when the market was wrong, as tested in Appendix~\ref{app:trading}.
Under the blind rule, every model loses money when it sells at the next state, significantly for three of the four, since prices tend to move away from a model that disagrees with them, and holding to settlement only shrinks the loss.
In the trading mode, Gemini keeps about half of its disagreement with the market, which improves its Brier score by 0.018 but leaves it 0.010 behind, and its trades make no significant profit.
These comparisons favor reporting intervals and the market price beside model scores.

\subsubsection{The Gap Diagnosis Points Mainly to Missing Information}
\label{sec:evidence-errors}

\begin{figure}[t]
\centering
\includegraphics[width=0.94\columnwidth]{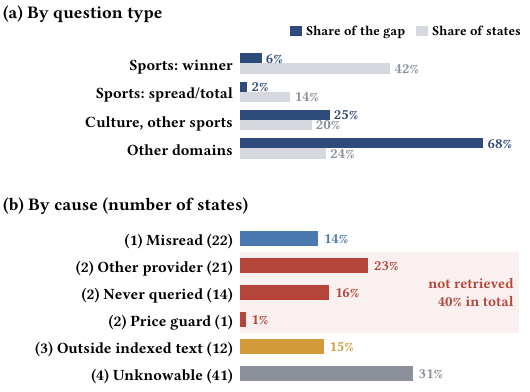}
\caption{Where the models' Brier gap to the market lies. Panel a shows shares of the gap by question type against each type's share of states; panel b shows causes across the 111 states trailing the market by more than 0.1, drawn from 49 questions.}
\label{fig:gap}
\end{figure}

\paragraph{The gap is concentrated outside standard sports contracts.}
Sports match winners, spreads, and totals make up more than half the states, and the models come within 0.004 Brier of the market on these questions.
Figure~\ref{fig:gap}a shows that almost all of the gap lies in other domains and culture questions, such as album sales, review scores, and awards.

\paragraph{The models trail in resolution, not in calibration.}
A Murphy decomposition separates calibration, whether events forecast at 70\% occur 70\% of the time, from resolution, the ability to give different probabilities to questions that resolve differently.
The mean of the four models is as well calibrated as the market and spreads its forecasts as widely, but its resolution is only about half the market's.
This shortfall in resolution is consistent with the missing-information diagnosis below.

\paragraph{The gap arises where the evidence is mixed or misleading.}
Evidence-direction labeling characterizes each state's evidence, as detailed in Appendix~\ref{app:direction}.
Where the evidence points toward the outcome, the models slightly beat the market, and only 12--17\% of each model's forecasts on the wrong side of 0.5 fall on such evidence, against 22\% for the market.
The whole gap thus arises where the evidence is mixed or points the wrong way, and it is concentrated: the 19\% of states in which the models trail the market by more than 0.1 Brier account for more than all of it.

\paragraph{Missing information costs more than misreading.}
For these states across all four question types, gap attribution assigns each state one of four causes in Figure~\ref{fig:gap}b.
\emph{Misread:} the models saw the fact but misweighted it, a small share.
Every model read, for instance, that the Panama Canal's caps allowed at most 226 transits in a week that needed more than 225, yet three put 0.43 to 0.60 on it, against the market's 0.06.
\emph{Not retrieved:} the fact was public but missing from the evidence, mostly because another provider returned it or the query-planning LLM never searched for it.
This is the largest cause.
\emph{Outside indexed text:} the signal lived in sales tracking, streaming counts, or data dashboards.
For one album, every provider returned projections of at most 46,000 first-week units, while the market priced more than 50,000 at 0.99, and it opened with 61,000.
\emph{Unknowable:} nothing public pointed to the outcome, and the market itself was unsure.
The judge identified no indexing-delay cases.

The judge attributes nearly three times as much of the gap in these states to missed public information as to misreading.
These diagnoses can guide query design for future live batches; RQ3 and RQ4 examine memory and the captured sources.

\subsection{Forecasts Do Not Sharpen, and Memory Steadies Them}
\label{sec:memory-results}

\begin{figure}[t]
\centering
\includegraphics[width=0.88\columnwidth]{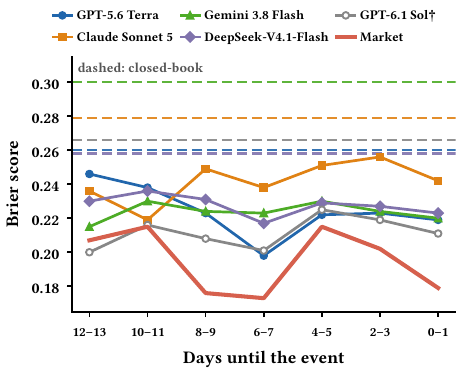}
\caption{Brier score by days until the event in the main condition with Tavily's evidence. Dashed: closed-book. $^\dagger$Released after the batch resolved.}
\label{fig:time}
\end{figure}

\paragraph{Forecasts do not sharpen as the event approaches.}
As Figure~\ref{fig:time} shows, the evidence keeps each model well below its closed-book score at every horizon, but the scores stay roughly flat as the event approaches, and the market leads every model throughout, consistent with the missing signals identified in Section~\ref{sec:evidence-errors}.
The models instead converge on one another, disagreeing less in the last days than a week before, as shown in Appendix~\ref{app:buckets}.
Every model also scores better on 10--14-day than on 2--5-day questions, by 0.057 for Gemini, though with only 21 short-horizon questions this may reflect their composition.

\begin{table}[t]
\caption{Memory variants of Gemini~3.8 Flash. Last: last-state Brier. Update: mean change between states. Agree: share in the market's direction. Ratio: size relative to the market's.}
\label{tab:stateful}
\centering
\footnotesize
\setlength{\tabcolsep}{3.5pt}
\begin{tabular}{@{}lccccc@{}}
\toprule
\textbf{Variant} & \textbf{TBS}$\downarrow$ & \textbf{Last}$\downarrow$ & \textbf{Update} & \textbf{Agree} & \textbf{Ratio} \\
\midrule
Main & \textbf{0.228} & \textbf{0.218} & 0.045 & 0.59 & 0.76 \\
Previous only & 0.232 & 0.221 & 0.049 & 0.59 & 0.83 \\
No memory & 0.237 & 0.230 & 0.084 & 0.50 & 1.24 \\
All so far & 0.232 & 0.227 & 0.059 & 0.49 & 0.91 \\
BM25 search & 0.241 & 0.249 & 0.075 & 0.46 & 1.10 \\
\bottomrule
\end{tabular}
\end{table}

\paragraph{Memory steadies forecasts more than it improves them.}
Table~\ref{tab:stateful} compares the main condition with the four memory variants of Table~\ref{tab:variants}, each of which changes only $M$ in Eq.~(\ref{eq:agent}).
Persistent notes give the best scores by small margins; only dropping all memory or using BM25 significantly worsens TBS, as shown in Appendix~\ref{app:memory}.
Without memory, however, the forecasts jump from state to state.
The mean update nearly doubles, exceeds the market's own moves, and goes in the market's direction only half the time.
With notes, updates are smaller than the market's and more often agree with it in direction, a mild under-reaction in line with earlier findings \citep{yuan2025evolvecast}.

\paragraph{Short notes reduce the cost of rereading.}
Rereading all evidence scores about as well as keeping notes but reads twice the input, 19.5M tokens per replay against 9.5M.
Searching the same pool with BM25 is the only alternative significantly worse in both metrics, and its forecasts differ from rereading by 0.057 on average, in line with the weakness of BM25 in controlled search benchmarks \citep{chen2026browsecompplus}.
Steadiness has a cost when the notes carry a wrong reading forward: for Norges Bank's September decision in Figure~\ref{fig:cases}a, all four models drifted down as their notes stressed softer inflation, while the market rose and the bank hiked.
Self-written notes offer a compact way to steady forecasts, but errors can persist, and none of the tested memory variants closes the gap to the market.

\subsection{Search-Source Differences Can Exceed Model Differences}
\label{sec:source-results}

\begin{figure}[t]
\centering
\includegraphics[width=0.94\columnwidth]{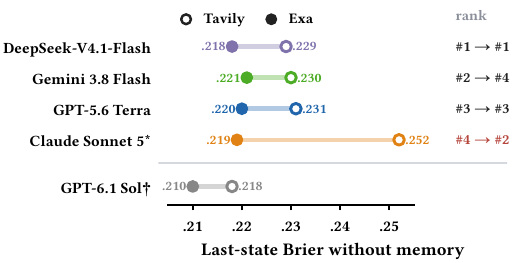}
\caption{Last-state Brier scores and model ranks without memory. Hollow markers: Tavily; filled markers: Exa. Kendall's $\tau=0$ across four models. $^*$Only significant change. $^\dagger$Not ranked.}
\label{fig:ranking}
\end{figure}

\begin{table}[t]
\caption{Gemini~3.8 Flash with each provider's evidence. The last two columns give the Brier score 0--1 and 4--7 days before the event. $^*$Worse than Tavily.}
\label{tab:source}
\centering
\footnotesize
\setlength{\tabcolsep}{4pt}
\begin{tabular}{@{}lccccc@{}}
\toprule
 & & & & \multicolumn{2}{c}{\textbf{Days to event}} \\
\cmidrule(l){5-6}
\textbf{Provider} & \textbf{Last-state}$\downarrow$ & \textbf{Accuracy}$\uparrow$ & \textbf{TBS}$\downarrow$ & 0--1 & 4--7 \\
\midrule
Tavily & 0.218 & 0.628 & \textbf{0.228} & 0.220 & \textbf{0.232} \\
Exa & 0.213 & \textbf{0.646} & 0.232 & 0.211 & 0.243 \\
Serper & 0.255$^*$ & 0.594 & 0.252$^*$ & 0.246 & 0.250 \\
Parallel & \textbf{0.212} & 0.639 & 0.229 & \textbf{0.209} & 0.246 \\
\bottomrule
\end{tabular}
\end{table}

\paragraph{The observed source spread exceeds the model spread.}
With the model fixed, switching the search API, the effect $\Delta_{k\to k'}$ of Eq.~(\ref{eq:effect}), moves Gemini's last-state Brier score by up to 0.043, more than the 0.027 that separates the four models in Table~\ref{tab:main}, as Table~\ref{tab:source} shows.
Most of this spread comes from Serper, significantly worse than Tavily, with the lowest share of relevant items in Appendix~\ref{app:providers}.
The effect includes retrieval, adaptive follow-ups, and evidence formatting under the shared collection policy.
Exa and Parallel stay close to Tavily, and none of the three leads at every horizon, consistent with the diagnosis in Section~\ref{sec:evidence-errors}, where another provider often held the missing fact.

\paragraph{The observed model order changes with the source.}
Figure~\ref{fig:ranking} shows different point-estimate rankings on Tavily's and Exa's evidence; with only four models, evidence for a systematic ranking effect is limited.
All four score better with Exa, but only Claude Sonnet~5 improves significantly, bringing its score close to the others.
Benchmarks should therefore fix or report each model's search source; pooling providers may help, as diverse evidence does in multi-agent forecasting \citep{li2026infodelphi}.

\section{Conclusion}

\fg provides a reusable protocol for prospectively recording search evidence and replaying it to compare forecasting models, memory strategies, and search sources.
The first batch shows that the tested date filters leave substantial leakage and that source differences can exceed model differences in accuracy.
All tested models trail the market, and our diagnosis attributes more of the gap in high-gap states to missed public information than to misreading.
Persistent notes steady forecasts with less input than rereading, but errors can carry forward as evidence changes.

These findings motivate improving both evidence acquisition and belief revision.
Frozen batches support controlled comparisons on recorded evidence, while new query policies require fresh live collection.
The shared collection protocol makes these studies repeatable on fresh questions as models and search policies evolve.

\bibliographystyle{ACM-Reference-Format}
\bibliography{references}

\appendix

\section{Batch Details}
\label{app:batch}
\label{app:scope}

Table~\ref{tab:stats} summarizes the first batch.
Selection balances horizon, domain, and activity within the two platforms' supply.
Candidate pools use bounded Polymarket sampling and exhaustive discovery within Kalshi's two-week window; their sports-heavy mix is consistent with real platform activity \citep{radde2026prediction}.
The quotas expand domain coverage without imposing equal domain shares, so the selected batch is grounded in the platforms' available questions.
Our 14-day batch contains 150 questions, compared with 80 in FutureX's contemporaneous weekly release for 16--22 September.\footnote{\url{https://huggingface.co/datasets/futurex-ai/Futurex-Online/commit/cd9efdbcf523f34f07f125b4a9b157a969bd6682}}
Quotas cover six domains, namely sports and culture, economics, politics, international affairs, business, and science and technology, and three levels of trading activity, the lower, middle, and upper thirds of 24-hour volume within each platform and domain.
The 576 states of the 144 settled questions are scored, and the 56 states with a failed search request are flagged.
The cancelled question was settled at 50/50, five questions are still open, and one question that its platform had not settled a week after the event was settled from the official result.

\begin{table}[!htb]
\caption{Statistics of the first batch. Document: a unique URL and content pair. Per-state and per-question counts are means.}
\label{tab:stats}
\centering
\small
\begin{tabular}{@{}lr@{}}
\toprule
\multicolumn{2}{@{}l}{\textit{Questions}} \\
Candidates screened / eligible & 6{,}754 / 279 \\
Candidates, Polymarket / Kalshi & 5{,}099 / 1{,}655 \\
Sports share, Polymarket / Kalshi & 95.5\% / 92.2\% \\
Selected (Kalshi / Polymarket) & 150 (104 / 46) \\
Horizon 2--5 / 6--9 / 10--14 days & 23 / 85 / 42 \\
Activity tier low / medium / high & 47 / 51 / 52 \\
Sports and culture / other domains & 109 / 41 \\
Settled to date (\% resolved Yes) & 144 (46.5\%) \\
Attribution questions / states & 49 / 111 \\
\midrule
\multicolumn{2}{@{}l}{\textit{Collection}} \\
Evidence states (per question) & 614 (2--7) \\
Search queries (shared first round) & 14{,}377 (9{,}816) \\
Search results & 131{,}542 \\
Unique URLs / documents & 47{,}581 / 102{,}419 \\
URLs per state: one provider / pooled & 39--45 / 133 \\
URLs per question, all states & 374 \\
Documents with page text & 68.8\% \\
Documents with publication date & 36.8\% \\
\bottomrule
\end{tabular}
\end{table}

\section{Implementation Details}
\label{app:config}

\subsection{Models}
\label{app:models}
Table~\ref{tab:models} lists each model's release date and reported training cutoff.
The four main models were released before the batch opened on 14 September 2026, and GPT-6.1 Sol only after it resolved.

\begin{table}[h]
\caption{Evaluated models. $^\dagger$Released after the batch resolved, with cutoff from third parties.}
\label{tab:models}
\centering
\footnotesize
\setlength{\tabcolsep}{6pt}
\begin{tabular}{@{}lll@{}}
\toprule
\textbf{Model} & \textbf{Released} & \textbf{Cutoff} \\
\midrule
GPT-5.6 Terra & 9 Jul 2026 & Feb 2026 \\
Claude Sonnet 5 & 30 Jun 2026 & Jan 2026 \\
Gemini 3.8 Flash & 2 Sep 2026 & Mar 2026 \\
DeepSeek-V4.1-Flash & 10 Sep 2026 & not disclosed \\
GPT-6.1 Sol$^\dagger$ & 29 Sep 2026 & Apr 2026 \\
\bottomrule
\end{tabular}
\end{table}

\subsection{Search Providers}
\label{app:providers}
Table~\ref{tab:providers} gives the outcome-free properties behind the choice of Tavily as the main provider.
Coverage is the query-planning LLM's rating, after the follow-up queries, of how well a provider's results cover the current-evidence, base-rate, and counterevidence purposes.
It leaves out resolution, whose evidence rarely exists before a question settles.
Relevance is the share of items whose title and snippet contain at least a third of the question's content words.
Models read text evidence from the search APIs without a separate full-page reading stage.
The market-price guard withholds 6.5\% of all results.
Collecting the batch's 14{,}377 requests cost about \$130 at these prices.

\begin{table}[h]
\caption{Search providers at the last state. Coverage rating, relevant and dated items, and list price per 1{,}000 requests.}
\label{tab:providers}
\centering
\footnotesize
\setlength{\tabcolsep}{5pt}
\begin{tabular}{@{}lcccr@{}}
\toprule
\textbf{Provider} & \textbf{Coverage} & \textbf{Relevant} & \textbf{Dated} & \textbf{Price} \\
\midrule
Tavily & \textbf{0.68} & \textbf{66\%} & 0\% & \$16 \\
Exa & 0.64 & 51\% & 39\% & \$15 \\
Serper & 0.53 & 42\% & 55\% & \$1 \\
Parallel & 0.64 & 56\% & 56\% & \$5 \\
\bottomrule
\end{tabular}
\end{table}

\subsection{Post-hoc Search Probe}
\label{app:leakage}

{\sloppy The probe draws its questions in proportion to the batch, 44 from sports and culture and 16 from other domains, and uses each provider's own date filter: Tavily's \texttt{end\_date}, Exa's \texttt{endPublishedDate}, and Serper's custom date range.
After the safeguards and deduplication, the post-hoc results give 12--16 items per question, against 15 in the live snapshot.\par}

\subsection{Trading Simulation}
\label{app:trading-sim}

Both trading experiments fill orders at half the question's quoted bid--ask spread, taken from its order book at collection and capped at 0.10, and pay Kalshi's taker fee of $0.07\,c(1-c)$ per contract priced $c$.
Contracts priced below 0.05 are not bought, since a single 1-cent contract would otherwise decide the total.
In the trading mode, the agent may stake up to \$1 per state, and a position still open after the last state is held to settlement.
The simulation allows fractional contracts and does not round Kalshi's fees up to the cent, so if anything it favors the trader.

\subsection{Diagnostic Labeling Tasks}
\label{app:labels}
Gemini~3.8 Flash performs both diagnostic tasks of Section~\ref{sec:evidence-errors}, with a separate prompt for each; the labels enter no score.
For evidence-direction labeling, it reads each state's evidence with the question, criteria, forecast time, and outcome, and rates how strongly the evidence points.
For gap attribution, it seeks a signal available beforehand rather than the outcome itself, and checks four pools in order: the evidence the model saw, Tavily items the market-price guard removed, the other providers' items, and later Tavily items dated at or before the state.
The first pool holding the fact sets the label, and otherwise the LLM judge picks one of the remaining causes of Figure~\ref{fig:gap}.

\section{Prompts}
\label{app:prompt}

Each forecast is one request consisting of instructions and a plain-text input, with a JSON schema requiring \path{probability}, \path{cited_evidence_ids}, and \path{rationale}.
The main condition also includes the note lists \path{evidence_notes}, \path{factors}, and \path{reasoning_notes}, capped at 8, 6, and 4 entries, respectively.
In the main condition the instructions are:

\begin{quote}\footnotesize\sloppy
You are forecasting the outcome of a binary question, which resolves according to the resolution criteria given with it. Give the probability that the first option is the outcome, as of the current time stated in the input. Use only the evidence provided. Each item shows its title, the website that published it, its publication date when known, and the date it was retrieved, followed by a snippet. Do not use any knowledge of events after the current time, including the outcome. In cited\_\allowbreak{}evidence\_\allowbreak{}ids, list the ids (such as E3) of the items you relied on. You also see your forecast and rationale from the previous state. Weigh them as one input among others, and do not anchor on them if the evidence now points elsewhere. You keep short notes on this question across states: the key evidence so far with its dates, the factors that have mattered, and how you have been updating. Revise them and return the updated lists in evidence\_\allowbreak{}notes, factors, and reasoning\_\allowbreak{}notes. The notes must stand on their own: do not refer to evidence ids, since items are renumbered at every state. Base any revision only on information available by the current time.
\end{quote}

\noindent The input gives the current time, the question, its options and resolution criteria, and the evidence items rendered as in Section~\ref{sec:setup} under the heading ``Evidence retrieved at the current state ($n$ items; earlier evidence appears only if it was retrieved again)''.
It ends with the previous forecast, its rationale, and the notes from earlier states.

At a question's first state the previous forecast is absent and the notes read ``none yet''.
Evidence identifiers that a model still writes into its notes are removed before the notes are carried forward.
The other conditions change the request as follows.
\emph{Previous only}: the notes, the note-keeping sentences, and the note lists in the schema are dropped.
\emph{No memory}: in addition, the previous forecast, its rationale, and the sentences about them are dropped.
\emph{All so far}: as no memory, with the evidence section headed ``Evidence ($n$ items, in the order they were first retrieved)''.
\emph{Closed-book}: no evidence or memory, and the sentences about evidence and citations are replaced by ``No evidence is provided: rely only on the question and your general background knowledge.''
\emph{BM25 search}: as no memory, but the model first receives only the current time, the question, and its resolution criteria, and is asked to ``propose up to 3 short, neutral web-search-style queries that would help resolve this forecasting question as of the current time, without presuming an outcome''.
The documents that BM25 retrieves for these queries from the evidence so far form the evidence section.

\section{Additional Results}
\label{app:results}

\subsection{Ensembles and Blends}
\label{app:ensemble}

Since LLM ensembles can rival human crowds \citep{schoenegger2024silicon} and a weighted blend of a model and the crowd can beat both \citep{halawi2024forecasting}, we score the mean $\bar p_t$ and median of the four models' forecasts and blends $w\bar p_t+(1-w)m_t$ with the market price $m_t$, for $w=0.5$ and for the $w=0.2$ of \citet{halawi2024forecasting}.
As Table~\ref{tab:app-ensemble} shows, the mean and median of the four models only match the best single model, at the last state and over all states.
Blending in the market helps, the more so the larger its weight, but even the blend that gives the market a weight of 0.8 falls just short of the market alone.

\begin{table}[h]
\caption{Brier score of model ensembles and model--market blends, where $w$ is the models' weight.}
\label{tab:app-ensemble}
\centering
\footnotesize
\begin{tabular}{@{}lcc@{}}
\toprule
\textbf{Forecaster} & \textbf{Last state}$\downarrow$ & \textbf{All states}$\downarrow$ \\
\midrule
Best single model & 0.217 & 0.222 \\
Mean of four & 0.219 & 0.224 \\
Median of four & 0.215 & 0.222 \\
Blend, $w=0.5$ & 0.189 & 0.205 \\
Blend, $w=0.2$ & 0.179 & 0.200 \\
\textit{Market} & \textbf{0.176} & \textbf{0.199} \\
\bottomrule
\end{tabular}
\end{table}

\subsection{Trading}
\label{app:trading}

Table~\ref{tab:trading} gives the profit of both trading experiments, under the rules of Appendix~\ref{app:trading-sim}.

\begin{table}[h]
\caption{Profit in cents per dollar staked, with 95\% intervals over questions.}
\label{tab:trading}
\centering
\footnotesize
\setlength{\tabcolsep}{3pt}
\begin{tabular}{@{}lrcc@{}}
\toprule
\textbf{Trader} & \textbf{Trades} & \textbf{Sold at next state} & \textbf{Held to settlement} \\
\midrule
\multicolumn{4}{@{}l}{\textit{Blind rule}} \\
GPT-5.6 Terra & 231 & $-13.8$ [$-23.4$, $-2.5$] & $-5.7$ [$-28.7$, $19.1$] \\
Claude Sonnet 5 & 278 & $-12.4$ [$-20.7$, $-3.6$] & $-3.7$ [$-23.7$, $18.8$] \\
Gemini 3.8 Flash & 214 & $-12.7$ [$-23.2$, $-0.4$] & $-3.8$ [$-28.4$, $25.2$] \\
DeepSeek-V4.1-Flash & 224 & $-10.9$ [$-20.8$, $0.8$] & $-5.4$ [$-28.1$, $20.4$] \\
\midrule
\multicolumn{4}{@{}l}{\textit{Trading mode, own exits}} \\
Gemini 3.8 Flash & 177 & \multicolumn{2}{c}{$-5.4$ [$-26.4$, $19.1$]} \\
\bottomrule
\end{tabular}
\end{table}

\begin{figure}[h]
\centering
\includegraphics{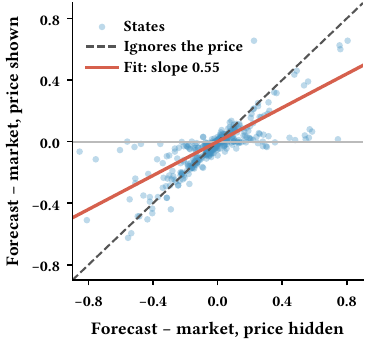}
\caption{Gemini hedges toward the market when shown its price. Forecast minus market price at the 575 scored states that have a market price, with the price hidden and shown. The dashed line marks ignoring the price, and the horizontal axis marks copying it.}
\label{fig:deference}
\end{figure}

Figure~\ref{fig:deference} shows how Gemini uses the price in the trading mode.
It keeps about half of its disagreement with the market and seldom copies the price outright.
Where its blind forecast differs clearly from the market, it almost always stays on the same side and only narrows the gap, so it hedges toward the market rather than defers to it.

\subsection{Evidence Direction and Properties}
\label{app:direction}

Table~\ref{tab:evidence} groups the states using the evidence-direction labeling task of Section~\ref{sec:evidence-errors}.
Where the evidence points toward the outcome, the models score about as well as the market or better, and few of their wrong-side forecasts fall there.
The gap opens where the evidence is mixed or points the other way.

\begin{table}[h]
\caption{Forecasts by the direction of a state's evidence (main condition, Tavily). The Brier columns use the LLM judge's labels. The wrong-side split counts weak support as mixed and requires at least moderate strength for toward. $n$: number of wrong-side forecasts.}
\label{tab:evidence}
\centering
\footnotesize
\setlength{\tabcolsep}{2pt}
\begin{tabular}{@{}lccc@{\hspace{8pt}}cccc@{}}
\toprule
& \multicolumn{3}{c}{\textbf{Brier}$\downarrow$} & \multicolumn{4}{c}{\textbf{Wrong side of 0.5}} \\
\cmidrule(r){2-4}\cmidrule(l){5-8}
\textbf{Forecaster} & \textbf{Toward} & \textbf{Mixed} & \textbf{Against} & \textbf{$n$} & \textbf{Toward} & \textbf{Mixed} & \textbf{Against} \\
\midrule
GPT-5.6 Terra & 0.133 & 0.266 & 0.387 & 216 & 17\% & 32\% & 50\% \\
Claude Sonnet 5 & 0.140 & 0.301 & 0.449 & 223 & 13\% & 36\% & 50\% \\
Gemini 3.8 Flash & 0.126 & 0.277 & 0.405 & 219 & 13\% & 34\% & 52\% \\
DeepSeek-V4.1-Flash & 0.120 & 0.273 & 0.429 & 206 & 12\% & 33\% & 54\% \\
\textit{Market} & 0.137 & 0.222 & 0.313 & 185 & 22\% & 29\% & 49\% \\
\midrule
States & 310 & 142 & 122 & & & & \\
\bottomrule
\end{tabular}
\end{table}

Simple evidence measures, such as the number of items, their lexical relevance, the share of new URLs, and the mix of query purposes, are at most weakly related to the Brier score, with no Spearman correlation above 0.11 in magnitude.
The size of the models' updates follows the market's moves, with a correlation of 0.26, but not the arrival of new URLs, with 0.02.

\subsection{Model Disagreement over Time}
\label{app:buckets}

\begin{figure}[tb]
\centering
\includegraphics{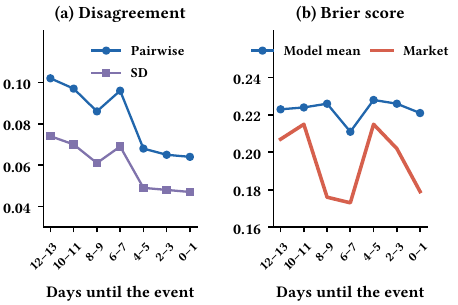}
\caption{Model disagreement and accuracy by days until the event. (a) Disagreement among the four models, as mean pairwise difference and standard deviation, and (b) Brier of their mean forecast and of the market.}
\label{fig:buckets}
\end{figure}

Figure~\ref{fig:buckets} shows that the models disagree more a week or more before the event than in its last days, while the Brier score of their mean forecast barely changes.
The buckets shrink from about 140 states in each two-day bucket of the last week to 18 two weeks out, so the earliest points are noisier.

\subsection{Memory Ablations}
\label{app:memory}

\begin{table}[tb]
\caption{Paired differences from the main condition (positive: worse). $\Delta$Last: last-state Brier. $G_t$: forecast minus the main condition's at the same state, with mean $\bar G$ and mean absolute value $|G|$.}
\label{tab:app-memory}
\centering
\footnotesize
\setlength{\tabcolsep}{2.5pt}
\resizebox{\columnwidth}{!}{%
\begin{tabular}{@{}lccccc@{}}
\toprule
\textbf{Condition} & $\Delta$\textbf{Last} [95\% CI] & $\Delta$\textbf{TBS} [95\% CI] & $\bar G$ & $|G|$ & $|G|>0.05$ \\
\midrule
Previous only & $+$0.003 [$-$0.004, 0.012] & $+$0.004 [$-$0.002, 0.009] & $+$0.003 & 0.025 & 13\% \\
No memory & $+$0.013 [$-$0.003, 0.032] & $+$0.010 [0.002, 0.018] & $+$0.008 & 0.042 & 21\% \\
All so far & $+$0.009 [$-$0.001, 0.022] & $+$0.004 [$-$0.003, 0.012] & $\pm$0.000 & 0.029 & 15\% \\
BM25 search & $+$0.031 [0.006, 0.058] & $+$0.013 [0.004, 0.023] & $+$0.012 & 0.059 & 28\% \\
\bottomrule
\end{tabular}}
\end{table}

Table~\ref{tab:app-memory} gives the paired differences of each RQ3 condition from the main condition and the per-state forecast difference $G_t$.
Searching the accumulated evidence with BM25 also scores worse at the last state than rereading the same pool, though not significantly.

\subsection{Case Studies}
\label{app:cases}

\begin{figure}[tb]
\centering
\includegraphics{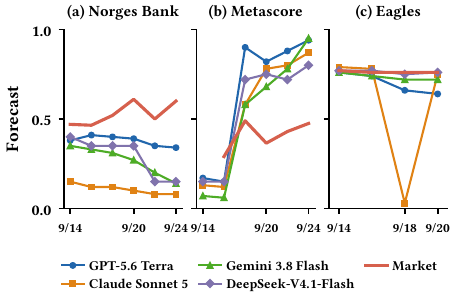}
\caption{Three case questions, all resolved Yes. Forecasts at each state (main condition, Tavily) and the market in red on (a) Norges Bank's hike, (b) a Metascore above 90, and (c) the Eagles beating the Titans.}
\Description{Three small line plots of the four models' forecasts and the market price over the evidence states of three questions.}
\label{fig:cases}
\end{figure}

Figure~\ref{fig:cases} traces three questions chosen by fixed rules: (a) the largest miss shared by all models while the market was right, (b) the largest update toward the outcome, and (c) the largest gap between two models' updates.
Together they mark the boundaries of what live evidence can do.
In (a), the notes kept stressing softer inflation, so memory held the models on a wrong reading while the market rose, as Section~\ref{sec:results} discusses.
In (b), Metascores above the threshold appeared the day after release, and the models rose sharply within one state while the market stayed below 0.5.
The decisive signal was published on the Web as soon as it existed, so retrieval put every model ahead of the market, the reverse of the usual gap.
In (c), three models read the same evidence correctly, so the error lies in one model's reading of undated pages rather than in retrieval, and it vanished at the next state.

\clearpage
\end{document}